\documentclass[conference]{IEEEtran}
\IEEEoverridecommandlockouts

\usepackage{cite}
\usepackage{amsmath,amssymb,amsfonts}

\usepackage{graphicx}
\usepackage{textcomp}
\usepackage{xcolor}
\usepackage{float}

\usepackage{graphicx}
\usepackage{hyperref}
\usepackage{algorithm}
\usepackage{algpseudocode}
\usepackage{array}
\usepackage{booktabs}
\usepackage{multirow}
\usepackage[compatibility=false]{caption}
\usepackage{subcaption}

\def\BibTeX{{\rm B\kern-.05em{\sc i\kern-.025em b}\kern-.08em
    T\kern-.1667em\lower.7ex\hbox{E}\kern-.125emX}}
\begin{document}

\title{Hybrid Framework for Robotic Manipulation:
 Integrating Reinforcement Learning and Large
 Language Models\\
{\footnotesize \textsuperscript{}
}
}

\author{\IEEEauthorblockN{Md Saad}

%Department of Mechanical Engineering, Jamia Millia Islamia, New Delhi, India
\IEEEauthorblockA{\textit{Department of Mechanical Engineering,} \\
\textit{Jamia Millia Islamia,}\\
New Delhi, India \\
mohammed.saadsalik@gmail.com}
\and

\IEEEauthorblockN{Sajjad Hussain}
\IEEEauthorblockA{\textit{School of Architecture,} \\
\textit{Technology and Engineering,} \\
\textit{University of Brighton,}\\
Brighton, UK \\
s.hussain4@brighton.ac.uk}
\and

\and
\IEEEauthorblockN{Mohd Suhaib}
\IEEEauthorblockA{\textit{Department of Mechanical Engineering,} \\
\textit{Jamia Millia Islamia}\\
New Delhi, India \\
msuhaib@jmi.ac.in}
}

\maketitle

\begin{abstract}
This paper introduces a new hybrid framework that combines Reinforcement Learning (RL) and Large Language Models (LLMs) to improve robotic manipulation tasks. By utilizing RL for accurate low-level control and LLMs for high-level task planning and understanding of natural language, the proposed framework effectively connects low-level execution with high-level reasoning in robotic systems. This integration allows robots to understand and carry out complex, human-like instructions while adapting to changing environments in real-time. The framework is tested in a PyBullet-based simulation environment using the Franka Emika Panda robotic arm, with various manipulation scenarios as benchmarks. The results show a 33.5\% decrease in task completion time and enhancements of 18.1\% and 36.4\% in accuracy and adaptability, respectively, when compared to systems that use only RL. These results underscore the potential of LLM-enhanced robotic systems for practical applications, making them more efficient, adaptable, and capable of interacting with humans. Future research will aim to explore sim-to-real transfer, scalability, and multi-robot systems to further broaden the framework’s applicability.
\end{abstract}


\section{Introduction}
In recent years, the field of robotics has seen significant growth, largely due to advancements in artificial intelligence (AI) and machine learning (ML) \cite{soori2023artificial}. One of the standout developments in this area is Reinforcement Learning (RL), which has proven highly effective in enabling robots to carry out low-level control tasks like manipulation and navigation with impressive precision and autonomy \cite{tang2024deep}. Concurrently, Large Language Models (LLMs) have transformed natural language understanding and task planning, providing capabilities for reasoning and decision-making that resemble human thought processes \cite{wang2024}. However, even with the progress made in these separate areas, the opportunity to merge RL and LLMs to develop more intelligent and adaptable robotic systems has not been fully explored \cite{jeong2024survey}.

This paper aims to fill that gap by introducing a hybrid framework that combines RL for executing low-level tasks with LLMs for high-level task planning and decision-making. This integration allows robots to comprehend and carry out complex, human-like instructions while adjusting to changing environments. By harnessing the advantages of both RL and LLMs, the proposed framework seeks to improve efficiency, accuracy, and adaptability in robotic manipulation tasks. The framework is evaluated in a PyBullet-based simulation environment using the Franka Emika Panda robotic arm, where various manipulation scenarios serve as benchmarks for assessing its performance.

\section{Related Work}
The integration of Reinforcement Learning (RL) and Large Language Models (LLMs) in robotics has become a hot topic in recent years. Gomes et al \cite{gomes2021} investigated deep RL for robotic pick-and-place tasks, achieving an impressive 84\% accuracy by utilizing MobileNet and depth cameras for robotic grasping in simulated environments. Shahid et al \cite{shahid2022} showcased the effectiveness of continuous control actions for robotic manipulation through the Proximal Policy Optimization (PPO) and Soft Actor-Critic (SAC) algorithms. Their research achieved a remarkable 100 \% success rate in grasping tasks with the Franka Emika Panda robot, successfully transferring learned behaviors from simulations to real-world scenarios.

Chu et al \cite{chu2023} introduced Lafite-RL, a framework that utilizes LLM feedback to speed up RL training. Their findings showed notable improvements in learning rates compared to traditional RL approaches. Likewise, Liu et al \cite{liu2024} proposed a self-corrected multimodal LLM framework for end-to-end robot manipulation, achieving 79 \% accuracy on known objects and 69 \% on unseen categories by incorporating an error rectification mechanism. In the realm of agricultural robotics, Che et al \cite{che2024}  highlighted the potential of merging generative AI and computer vision for intelligent robotic systems, surpassing conventional methods in agricultural picking tasks despite facing environmental challenges. Other researchers have also explored the wider applications of LLMs in robotics. Wang et al \cite{wang2024} provided a thorough review of LLMs in robotics, emphasizing the opportunities and challenges that come with their integration into robotic systems. Kwon et al \cite{kwon2024} demonstrated the practicality of using LLMs as zero-shot trajectory generators, illustrating that LLMs can create low-level trajectories for robotic manipulation using straightforward prompts. Mower et al \cite{mower2024} introduced ROS-LLM, a framework that links LLMs with the Robot Operating System (ROS), making robot programming more accessible for non-experts through natural language.

Recent advances in robotics emphasize combining high-level semantic reasoning with low-level control for improved adaptability. Reinforcement Learning (RL) has shown success in control tasks but struggles with interpreting complex instructions. This gap is addressed by integrating Large Language Models (LLMs), which enable robots to understand and generalize tasks through language \cite{kim2024survey}. Jain et al. \cite{jain2023grounding} introduced dynamic reward specification via natural language. Huang et al. \cite{huang2023innermonologue} proposed InnerMonologue for internal task reasoning. Ahn et al. \cite{ahn2022saycan} developed SayCan, grounding LLM plans through feasibility scoring. Jiang et al. \cite{jiang2023llmplanner} presented LLM-Planner to generate behavior trees with commonsense reasoning. Khandelwal et al. \cite{khandelwal2023promptirl} introduced PromptIRL, using language prompts for inferring reward functions from demonstrations. These approaches highlight the potential of LLMs to enhance RL-based robotic manipulation through improved planning and interpretability.

%%%%%%%%%%%%%%%%%%%%%%%%%%%%%%%%%%%%

\section{Proposed Work}
In this work, we present a hybrid framework that integrates Reinforcement Learning (RL) and Large Language Models (LLMs) to achieve robust and adaptive robotic manipulation. The proposed framework leverages the planning and reasoning capabilities of LLMs for high-level task understanding and decomposition, while RL ensures precise low-level task execution.

\subsection{Framework Overview}
The framework consists of three primary components:
\begin{itemize}
    \item \textbf{Task Planner (LLM-based)}: The Task Planner interprets natural language instructions and decomposes them into a sequence of subtasks.
    \item \textbf{Skill Executor (RL-based)}: This component executes each subtask in the simulation environment, relying on trained RL policies for effective manipulation.
    \item \textbf{Simulation Environment}: The PyBullet simulation environment provides a realistic testing platform for evaluating the hybrid framework. It includes the Franka Emika Panda robotic arm, simulated sensory inputs, and various objects for manipulation.
\end{itemize}

\begin{figure}[H]
    \centering
    \includegraphics[width=0.55\textwidth]{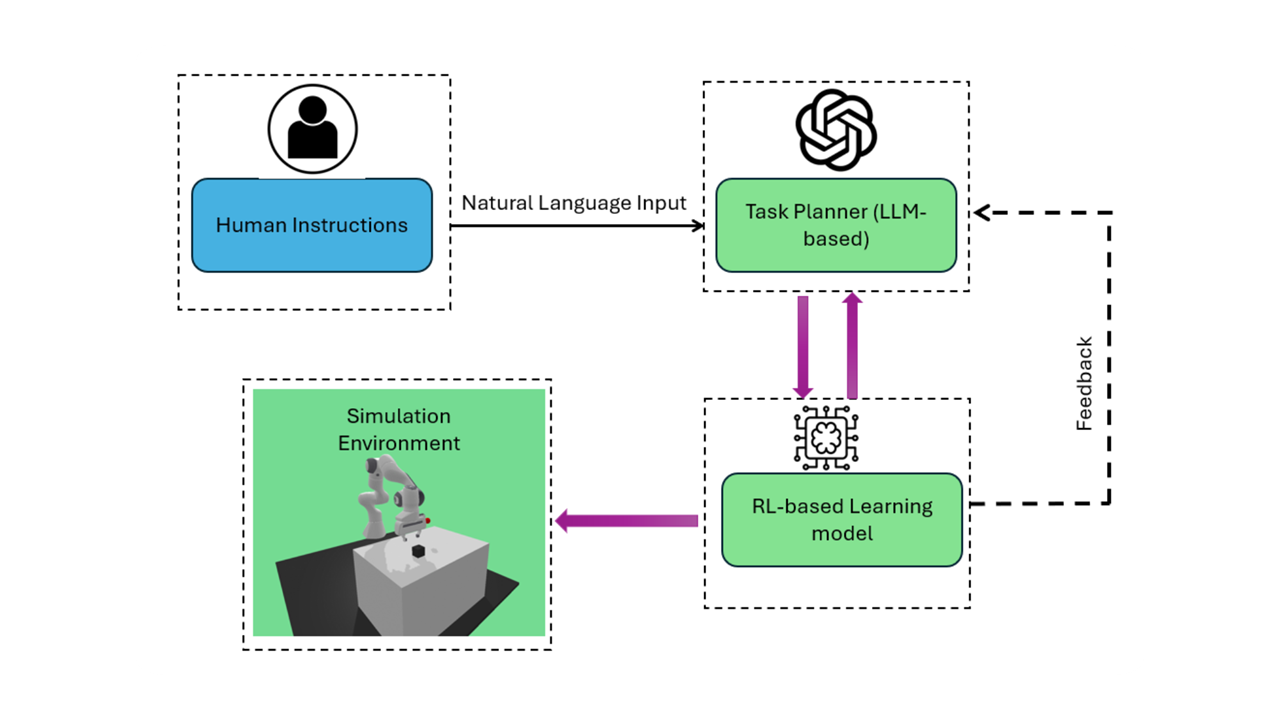}
    \caption{System Architecture: Integration of LLM and RL for Robotic Manipulation}
    \label{fig:architecture}
\end{figure}

\subsection{Workflow Description}
The overall workflow is as follows:
\begin{enumerate}
    \item \textbf{System Initialization}: The system initializes the LLM, RL agent, and simulation environment.
    \item \textbf{Natural Language Input}: The user provides high-level instructions, which are processed by the LLM.
    \item \textbf{Task Decomposition and Execution}: The LLM decomposes the instruction into subtasks, which are sequentially executed by the RL agent.
    \item \textbf{Feedback and Adaptation}: The LLM continuously monitors the environment for changes and updates the task plan when necessary.
\end{enumerate}

\subsection{Simulation Environment}
The PyBullet-based simulation environment is designed to evaluate the proposed framework in various object manipulation tasks, such as pick-and-place and obstacle avoidance. The environment includes the Franka Emika Panda robot and objects with varying properties.

\begin{figure}[H]
    \centering
    \includegraphics[width=0.5\textwidth]{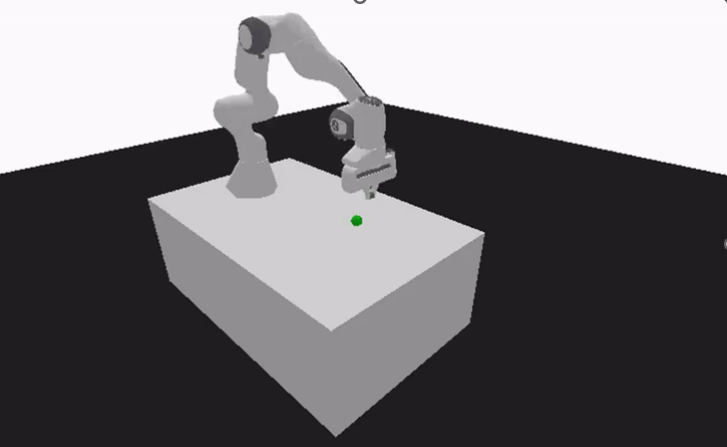}
    \caption{Simulation Environment: Franka Emika Panda Robot in PyBullet}
    \label{fig:simulation}
\end{figure}

\subsection{Integration of LLM and RL}
The Task Planner (LLM) and Skill Executor (RL) communicate through an integration layer. The LLM module provides high-level task decomposition, while the RL module ensures precise execution of each subtask. A feedback loop between the two modules enables dynamic task adaptation and error recovery.

%%%%%%%%%%%%%%%%%%%%%%%%%%%%%%%%%%%%%%

\section{Methodology}

\subsection{Simulation Environment}
The proposed framework is developed and tested in a simulation environment using \textbf{PyBullet}, a physics-based simulation platform. The simulation setup includes the \textbf{Franka Emika Panda} robotic arm, which is widely used for manipulation tasks due to its high precision and flexibility. The environment is designed to include various object manipulation scenarios such as pick-and-place tasks, object sorting, and dynamic obstacle avoidance. These tasks serve as benchmarks for evaluating the performance of the proposed framework.

The task setup consists of:
\begin{itemize}
    \item A \textbf{workspace} with predefined positions for objects and target locations.
    \item \textbf{Sensory inputs} simulated using depth cameras and force sensors.
    \item \textbf{Control strategies} implemented using reinforcement learning (RL) policies for executing low-level movements.
\end{itemize}
This simulation setup provides a controlled environment for training and testing the integration of reinforcement learning and large language models.

\subsection{Hybrid Framework}
\begin{figure}[h]
    \centering
    \includegraphics[width=0.5\textwidth]{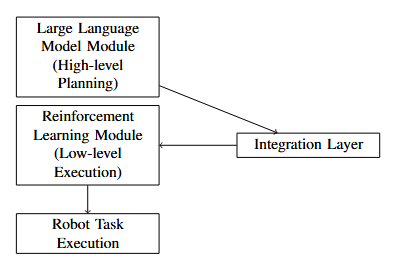}
    \caption{Hybrid Framework Integrating RL and LLMs}
    \label{fig:hybrid_framework}
\end{figure}

The hybrid framework integrates \textbf{Reinforcement Learning (RL)} for low-level task execution and \textbf{Large Language Models (LLMs)} for high-level task planning. The overall structure of this framework is depicted in Figure~\ref{fig:hybrid_framework}, illustrating the interaction between these components.

\subsubsection{Reinforcement Learning Module}
The reinforcement learning module is responsible for executing the low-level control policies required for robot task execution. In this work, we use \textbf{Proximal Policy Optimization (PPO)} and \textbf{Soft Actor-Critic (SAC)}, two widely used RL algorithms, to train policies for robot manipulation tasks.

The RL module operates as follows:
\begin{itemize}
    \item \textbf{State representation:} The robot perceives the environment through sensory inputs, including joint positions, velocities, and object positions.
    \item \textbf{Action selection:} Based on the learned policy, the RL agent selects the optimal action to manipulate objects.
    \item \textbf{Reward function:} A custom reward function is designed to encourage efficient and accurate task execution.
    \item \textbf{Training process:} The agent undergoes episodic training, continuously refining its policy based on interaction with the environment.
\end{itemize}
This module ensures that the robot can autonomously perform physical interactions necessary for completing the assigned tasks.

\subsubsection{Large Language Model Module}
The Large Language Model (LLM) module is responsible for understanding and interpreting high-level task instructions. We leverage models such as \textbf{GPT-based architectures} to process human-like commands and translate them into structured subtask sequences.

\textbf{Key functionalities:}
\begin{itemize}
    \item \textbf{Task interpretation:} LLMs parse human instructions (e.g., "Pick up the red cube and place it on the blue platform").
    \item \textbf{Task decomposition:} The high-level instruction is broken down into smaller subtasks, such as "Move to the cube," "Grasp the cube," and "Move to the target location."
    \item \textbf{Output generation:} The LLM generates structured command sequences, which are then fed into the RL module for execution.
\end{itemize}

This integration enables robots to follow complex, natural language-based instructions, making them more adaptable to human commands.

\subsubsection{Integration}
The integration of the RL and LLM modules is facilitated through an \textbf{Integration Layer}, which serves as the communication bridge between high-level planning and low-level execution.

\subsection*{Pseudocode: LLM and RL Integration for Robotic Manipulation}

\begin{algorithm}
\caption{LLM and RL Integration for Robotic Manipulation}
\begin{algorithmic}[1]
\State Initialize LLM and RL agent
\State Initialize simulation environment (PyBullet)
\While{task is not complete}
    \State Parse NLI (Natural Language Instruction) using LLM to generate subtasks
    \ForAll{subtask}
        \State Query RL agent to execute the subtask
        \State Monitor environment state
        \If{environment changes}
            \State Update task plan via LLM
        \EndIf
    \EndFor
\EndWhile
\State Return completion status
\end{algorithmic}
\end{algorithm}

This integration allows the system to operate in a dynamic manner, where the LLM provides reasoning and planning capabilities, while RL ensures precise execution.

\section{Results and Discussion}

This section presents the results of the proposed hybrid framework integrating Reinforcement Learning (RL) with Large Language Models (LLMs) for robotic manipulation. The evaluation was conducted in a PyBullet simulation environment using the Franka Emika Panda robot. The results highlight improvements in task completion time, accuracy, and adaptability compared to an RL-only approach.

\subsection{Task Completion Time}

\begin{figure}[H]
    \centering
    \includegraphics[width=0.5\textwidth]{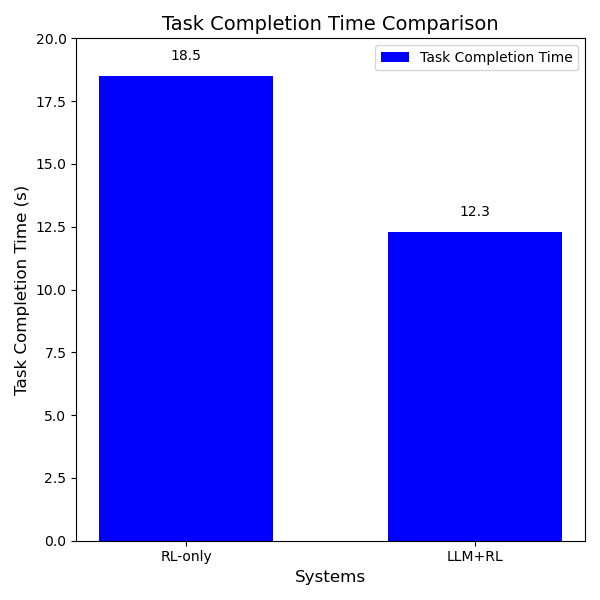}
    \caption{Task Completion Time Comparison}
    \label{fig:task_completion_time}
\end{figure}

Figure~\ref{fig:task_completion_time} illustrates the comparison of task completion time between the RL-only and LLM+RL systems. The RL-only system took an average of 18.5 seconds to complete tasks, while the LLM+RL system reduced this time to 12.3 seconds, representing a 33.5\% improvement in efficiency.

\subsection{Accuracy and Adaptability}

\begin{figure}[H]
    \centering
    \includegraphics[width=0.5\textwidth]{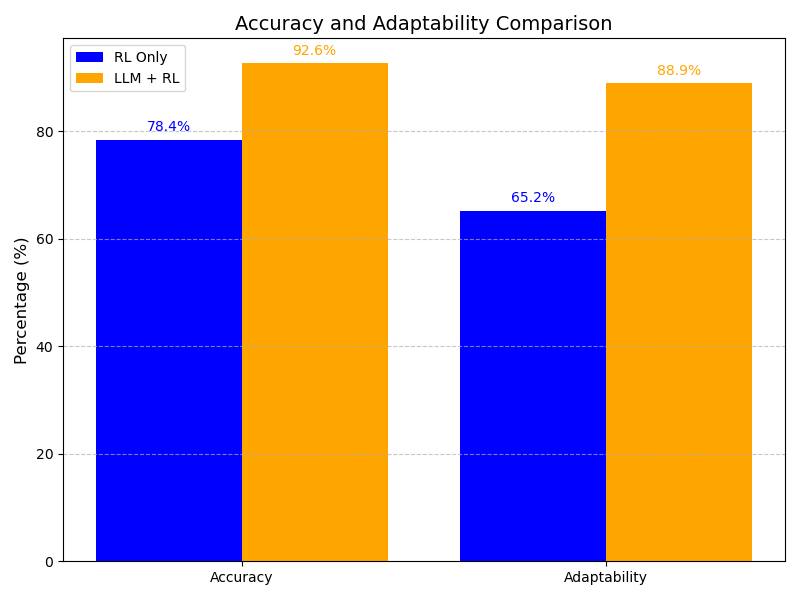}
    \caption{Accuracy and Adaptability Comparison}
    \label{fig:accuracy_adaptability}
\end{figure}

The proposed hybrid system also demonstrates significant improvements in accuracy and adaptability, as shown in Figure~\ref{fig:accuracy_adaptability}. The LLM+RL system achieved an accuracy of 92.6\% compared to 78.4\% for the RL-only system. Similarly, the adaptability increased from 65.2\% in the RL-only system to 88.9\% in the LLM+RL system, highlighting the robustness of the proposed approach in dynamic environments.

\subsection{Task Completion Time Across Experiments}

\begin{figure}[H]
    \centering
    \includegraphics[width=0.5\textwidth]{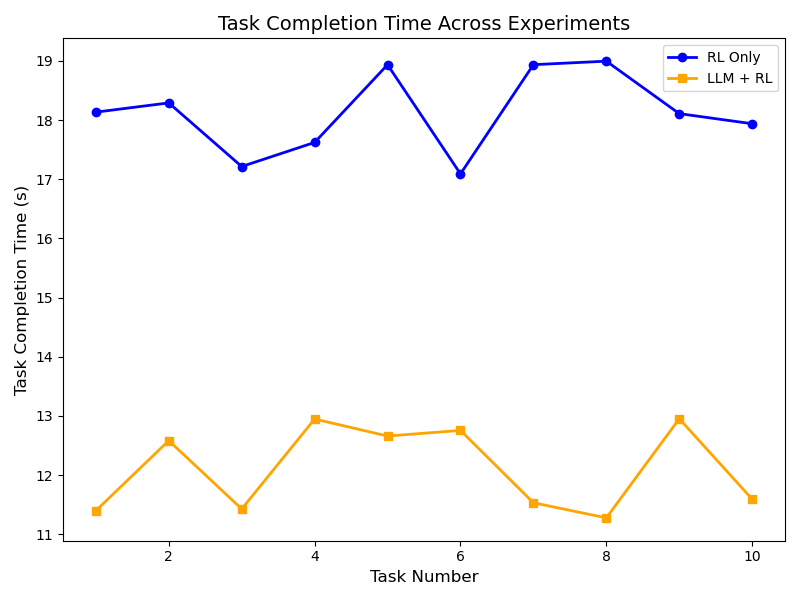}
    \caption{Task Completion Time Across Experiments}
    \label{fig:task_time_experiments}
\end{figure}

Figure~\ref{fig:task_time_experiments} shows task completion time across 10 different experimental runs. The RL-only system exhibited higher completion times with fluctuations, ranging from 17.5 to 19 seconds. In contrast, the LLM+RL system maintained a consistent average of approximately 12 seconds, demonstrating better stability and efficiency.

\subsection{Learning Progress}

\begin{figure}[H]
    \centering
    \includegraphics[width=0.5\textwidth]{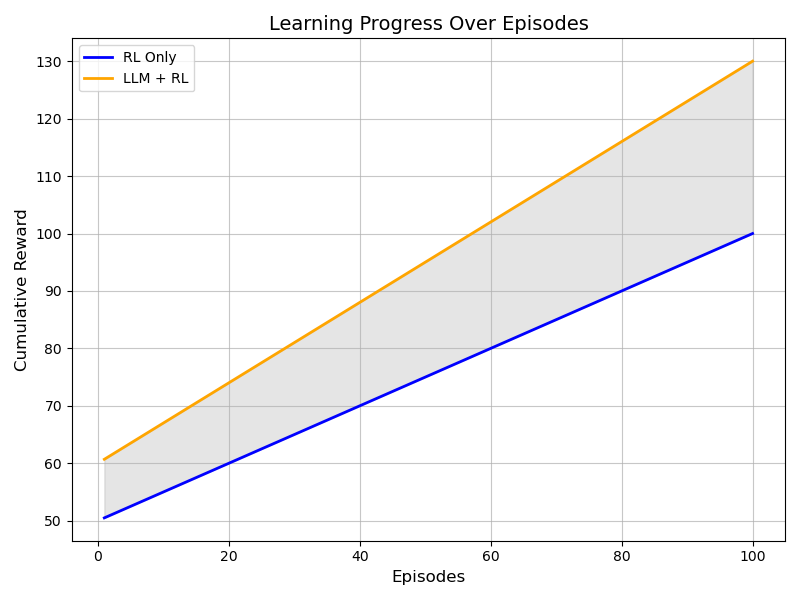}
    \caption{Learning Progress Over Episodes}
    \label{fig:learning_progress}
\end{figure}

The learning progress over 100 episodes is depicted in Figure~\ref{fig:learning_progress}. The cumulative reward for the LLM+RL system grew steadily and outperformed the RL-only system, reflecting the enhanced learning capability provided by the hybrid framework.

\subsection{Performance Comparison}

Table~\ref{table:performance_comparison} summarizes the performance metrics for the RL-only and LLM+RL systems. The LLM+RL system demonstrates notable improvements in task completion time, accuracy, and adaptability, reinforcing the effectiveness of integrating high-level task planning from LLMs with low-level execution from RL.

\begin{table}[H]
    \centering
    \renewcommand{\arraystretch}{1.3}
    \caption{Performance Comparison of RL-Only and LLM+RL Systems}
    \begin{tabular}{|p{2cm}|c|c|c|}
        \hline
        \textbf{Metric} & \textbf{RL Only} & \textbf{LLM + RL} & \textbf{Improvement (\%)} \\ \hline
        Task Completion Time (s) & 18.5 & 12.3 & 33.5 \\ \hline
        Accuracy (\%) & 78.4 & 92.6 & 18.1 \\ \hline
        Adaptability (\%) & 65.2 & 88.9 & 36.4 \\ \hline
    \end{tabular}
    \label{table:performance_comparison}
\end{table}

The results clearly demonstrate that integrating LLMs into robotic manipulation tasks significantly improves performance across multiple metrics. The hybrid framework's ability to interpret high-level natural language instructions and adapt to dynamic environments makes it more robust and efficient than traditional RL-based approaches. The feedback mechanism between the LLM and RL components ensures continuous task refinement, resulting in improved accuracy and adaptability. These findings highlight the potential for broader applications of LLM+RL systems in real-world robotic scenarios.

\section{Conclusion}

This paper presents a hybrid framework that combines Reinforcement Learning (RL) with Large Language Models (LLMs) to enhance robotic manipulation tasks. Results from PyBullet simulations using the Franka Emika Panda robot indicate notable improvements in task completion time, accuracy, and adaptability when compared to systems relying solely on RL. The LLM+RL approach achieves a 33.5 \% reduction in task completion time, while also boosting accuracy and adaptability by 18.1\% and 36.4\%, respectively. This integration allows for dynamic task adaptation and the ability to process natural language instructions, thereby making robotic systems more efficient and resilient. Future research will aim at real-world implementation, scalability for multiple robots, and improving sim-to-real transfer to broaden the framework’s applicability.

\bibliographystyle{IEEEtran}
\bibliography{ref}

\end{document}